\documentclass[letterpaper, 10 pt, conference]{ieeeconf}  

\IEEEoverridecommandlockouts                              

\usepackage{amsmath, mathtools}
\usepackage{xcolor}
\usepackage{tabularx}
\usepackage{multirow}
\usepackage{amsfonts}
\usepackage{gensymb}
\usepackage{comment}
\usepackage{textcomp}
\usepackage{rotating}

\newcommand\norm[1]{\left\lVert#1\right\rVert}

\title{\LARGE \bf
\textcolor{black}{Helical Tendon-Driven Continuum Robot with \\ Programmable Follow-the-Leader Operation}
}

\author{Behnam Moradkhani$^{1}$, Raghav Sankaranarayanan$^{2}$, Pejman Kheradmand$^{1}$, Harshith Jella$^{1}$,\\ Nicholas Ahn$^{3}$, Ajmal Zemmar$^{4}$ and Yash Chitalia$^{1}$
\thanks{*This work was supported in part by the Kentucky Spinal Cord and Head Injury Research Trust (KSCHIRT) Fund.}
\thanks{$^{1}$B. Moradkhani, P. Kheradmand, H. Jella, and Y. Chitalia are with the Healthcare Robotics and Telesurgery Laboratory (Heartlab), University of Louisville, Louisville, KY, USA.}%
\thanks{$^{2}$At the time this work was conducted, R. Sankaranarayanan was with the Healthcare Robotics and Telesurgery Laboratory (Heartlab), University of Louisville, Louisville, KY, USA.}%
\thanks{$^{3}$N. Ahn is with the Department of Orthopedic Surgery, University of Louisville School of Medicine, Louisville, KY, USA.}%
\thanks{$^{4}$A. Zemmar is with the Robley Rex Va Medical Center, Louisville VA Medical Center, Louisville, KY, USA.}%
\thanks{A. Zemmar and Y. Chitalia contributed equally to senior authorship.}%
\thanks{Corresponding Author: B. Moradkhani ({\tt \small b0mora01@louisville.edu})}}%

\begin{document}

\maketitle
\thispagestyle{empty}
\pagestyle{empty}

\begin{abstract}
Spinal cord stimulation (SCS) is widely used for pain management and has recently shown promise in promoting functional recovery following spinal cord injury. Effective activation of motor neurons requires accurate placement of SCS leads in the ventral or lateral epidural space, where corticospinal and rubrospinal motor fibers are located, a task that remains challenging with conventional manual steering. This study presents a static modeling framework for ExoNav, a steerable robotic tool designed to enable precise navigation to ventral and lateral epidural targets. A Cosserat rod formulation is employed to relate tendon actuation forces to the robot shape, and a simulation environment is developed based on this model, incorporating gravity as an external load. Experimental validation across four prototypes yields tip position root mean square errors (RMSEs) of 1.76 mm, 2.33 mm, 2.18 mm, and 1.33 mm, corresponding to 2.3\%, 3.1\%, 3.4\%, and 2.1\% of the robot length, respectively. Owing to its helical deformation under actuation, ExoNav enables follow-the-leader (FTL) motion when combined with insertion and rotation degrees of freedom, which is demonstrated in both simulation and experiments. The proposed simulation computes optimal tendon tensions to track desired FTL paths under gravity-induced deformation. Three FTL experiments show repeatable end-effector alignment with a maximum RMSE of 3.75 mm (less than 5\% of the robot length). Finally, phantom experiments demonstrate successful teleoperated navigation to ventral and lateral spinal cord targets, as well as the dorsal root ganglia, highlighting ExoNav’s potential for both motor function recovery and pain management.
\end{abstract}
\section{INTRODUCTION}
\begin{figure}
    \centering
    \includegraphics[width=0.75\linewidth]{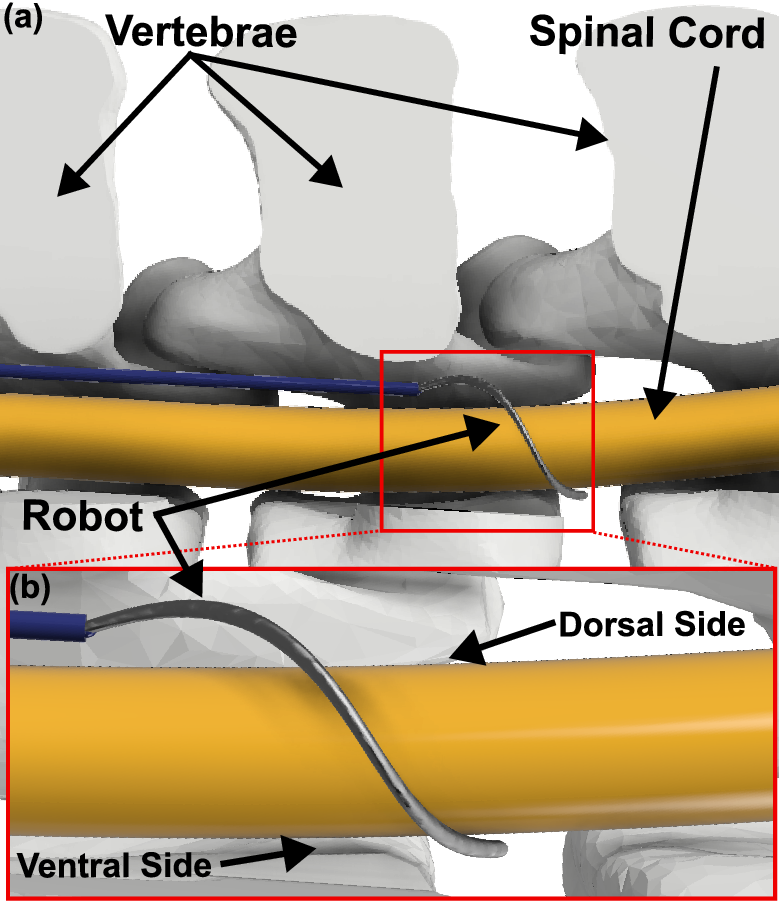}
    \caption{\textcolor{black}{(a) cross-sectional view of spinal vertebral segments and the spinal cord with ExoNav deformed in a helical shape around the spinal cord. (b) Close view of ExoNav, navigated from the dorsal spinal cord (entry point) to the ventral spinal cord (target region).}}
    \label{fig:Intro}
\end{figure}
Spinal cord stimulation (SCS) is a neuromodulation intervention to treat drug-refractory pain \cite{caylor2019spinal}. SCS has also demonstrated versatility in recovery of post spinal cord injury, including the restoration of both voluntary and involuntary functions \cite{afridi2024ventral, darrow2019epidural, gill2018neuromodulation}. Currently, the SCS leads are placed in the dorsal epidural space since this has been well established for pain treatment. While the pain fibers are located dorsally (along thoracic and cervical fibers), motor fibers are located in the ventral and lateral spinal cord. The current standard of manual steering does not permit the surgeon to place the SCS leads in the ventral and lateral epidural space although the benefits and efficacy have been demonstrated \cite{afridi2024ventral}.

Despite significant advancements in surgical robotics, limited attention has been given to developing robotic solutions specifically for steering SCS leads. While a system \cite{torlakcik2021magnetically} has been proposed, it remains a conceptual perspective. More recently, a hybrid tendon- and magnet-actuated continuum robot \cite{jella2024exonav} \textcolor{black}{was} designed to navigate the dorsal spinal cord. However, \textcolor{black}{this} system featured a relatively large form factor and lacked evidence of safe navigation in proximity to the delicate spinal structures. In this context, one of the most promising approaches for achieving safe and minimally invasive navigation, particularly relevant to continuum robotics, is the follow-the-leader (FTL) motion.

FTL motion has evolved into a significant research robotic challenge, receiving heightened attention over the past ten years \cite{culmone2021follow}. This motion strategy entails the device’s centerline tracking the path of its leading tip along a predetermined curved trajectory, thereby minimizing disruption to the surrounding environment. A variety of designs have been developed to achieve FTL motion. These include shape memory alloy (SMA) robots \cite{palmer2014real}, soft inflatable robots \cite{hawkes2017soft, coad2020retraction}\textcolor{black}{, concentric tube robots \cite{gilbert2015concentric, garriga2018complete, xie2025curvature}}, and tendon-driven continuum robots \cite{jeong2020design, neumann2016considerations}\textcolor{black}{, \cite{chen2014modular}}. FTL motion proves especially advantageous in surgical contexts, where navigating delicate anatomical regions, such as the spinal cord, is crucial. Continuum surgical robots enable minimally invasive procedures by accessing complex internal pathways, improving precision in surgeries like laparoscopy and endoscopy, while also reducing patient recovery times and streamlining surgical processes \cite{da2020challenges, omisore2020review}. Consequently, numerous studies have focused on integrating FTL motion into continuum robotic systems to enhance accuracy and minimize risks in sensitive surgical applications.

Geometry-based modeling approaches are commonly employed for tendon-driven continuum robots and catheters due to their ease of derivation, low computational cost, and suitability for real-time applications \cite{webster2010design, hachen2022model}. However, these methods often rely on simplifying assumptions such as piecewise constant curvature and inextensibility, which introduce significant errors \cite{trivedi2008soft, burgner2015continuum}, \textcolor{black}{\cite{moradkhani2025exonav}}. In surgical settings where precise control and accuracy are critical, such approximations may prove inadequate. As a result, the field is increasingly shifting toward mechanics-based models \textcolor{black}{\cite{rucker2011statics, wenlong2013mechanics, swaney2017design}, especially} those grounded in Cosserat rod theory \cite{rucker2011statics}, which offer a more accurate representation of continuum robot behavior by accounting for axial strain, shear, and tendon-induced effects \cite{dupont2010design}. With growing computational capabilities, these sophisticated models are becoming viable for surgical applications, enabling higher fidelity in both planning and control.

\textcolor{black}{In this paper, we present a helically notched tendon-driven continuum robot for navigating probes to lateral and ventral spinal cord fiber tracts and develop a Cosserat rod model to describe its mechanics (Sections \ref{subsec:design_and_model_1} and \ref{subsec:design_and_model_2}). Tendon actuation generates an adjustable helical shape whose parameters can be selected according to the target spinal segment and epidural space geometry. The helical configuration also enables FTL motion under appropriate translation and rotation, as demonstrated in \cite{ha2019pediatric}. These motion inputs and tendon tensions are programmed and optimized based on the developed Cosserat rod model (Sections \ref{subsec:design_and_model_3} and \ref{subsec:design_and_model_4}). Finally, the model and FTL motion are experimentally validated in a phantom spinal cord model (Section \ref{sec:simulation_and_experiment}), followed by conclusions and future research directions (Section \ref{sec:conclusion}).}


\section{\textcolor{black}{DESIGN AND MODEL}} \label{sec:design_and_model}

\begin{figure}[b]
    \centering
    \includegraphics[width=1.0\linewidth]{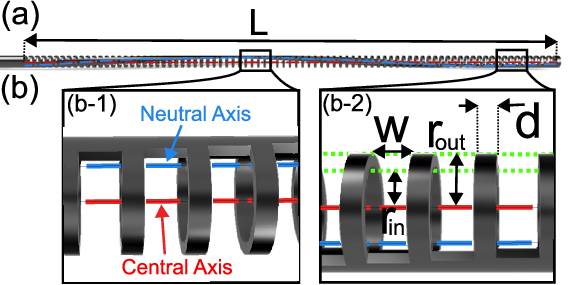}
    \caption{(a) Structure of \textit{ExoNav}: a nitinol tube with helically machined rectangular notches along its length. \textcolor{black}{Straight central axis and helical neutral axis are depicted over the structure.} (b) Zoomed-in view\textcolor{black}{s} showing the \textcolor{black}{(a) central and neutral axis locations, and (b)} notch pattern\textcolor{black}{s} and \textcolor{black}{their} dimensions.}
    \label{fig:robot_structure}
\end{figure}

\textcolor{black}{\textit{ExoNav} is a micromachined nitinol tube robot designed for precise placement of SCS electrodes within the lateral and ventral epidural space. The robot consists of a hollow nitinol tube with inner and outer radii of $r_{in}$ and $r_{out}$, featuring helical unidirectional asymmetric notches \cite{chitalia2018design}. Each notch has a width $w$ and is separated by a distance $d$ along the tube axis, with the notch pattern completing one full circumferential rotation over the robot length $L$ (Fig. \ref{fig:robot_structure}). A nitinol tendon of radius $r_t$, routed through the tube and anchored at the tip, deforms the robot into a helical shape (Fig. \ref{fig:cross-section}(a)). The notched tube is enclosed within an outer tube that constrains a variable proximal portion of the robot. Four prototypes were fabricated with different tube diameters ($r_{in}$ and $r_{out}$) and notch depths ($\psi$), resulting in varying sizes and compliances (Table \ref{tab:prototype_specs}). These variations were used to evaluate the proposed theoretical model while accounting for the dimensional constraints of the intended application.}

\begin{table} [t]
\caption{Prototypes physical specifications.}
\label{tab:prototype_specs}
\begin{center}
\begin{tabular}{|c|c|c|c|c|}
\hline
  prototype & 1 & 2 & 3 & 4 \\
  \hline
  $r_{in}$~(mm) & 0.457 & 0.457 & 0.851 & 0.851 \\

  $r_{out}$~(mm) & 0.572 & 0.572 & 0.953 & 0.953 \\

  $w$~(mm) & 0.4 & 0.4 & 0.5 & 0.5 \\

  $d$~(mm) & 0.2 & 0.2 & 0.3 & 0.3 \\

  $L$~(mm) & 75.4 & 75.4 & 63.27 & 63.27 \\

  $r_t$~(mm) & 0.05 & 0.05 & 0.1 & 0.1 \\

  $\psi~(\degree)$ & 90 & 107 & 90 & 126 \\
\hline
\end{tabular}
\end{center}
\end{table}

\subsection{Static Equilibrium Equations for Tendon-Driven Robots} \label{subsec:design_and_model_1}

Our proposed model is based on the general Cosserat rod theory for tendon-driven continuum robots  \cite{rucker2011statics}. All the kinematic parameters and mechanical properties, varying over the robot length, are defined as functions of the arc-length $s$, and $d/ds$ is denoted by the \textcolor{black}{dot ( }$\dot{}$ \textcolor{black}{)} notation.

\begin{equation} \label{p_dot}
    \textcolor{black}{\dot{\textbf{p}} = \textbf{Rv}}
\end{equation}
\begin{equation} \label{R_dot}
    \textcolor{black}{\dot{\textbf{R}} = \textbf{R}\hat{\textbf{u}}}
\end{equation}

$\textcolor{black}{\textbf{p}} \in \mathbb{R}^3$ is the position vector of the robot \textcolor{black}{neutral axis} defined in the global frame (see Fig. \ref{fig:cross-section}(a)) and $\textcolor{black}{\textbf{R}} \in SO(3)$ describes the robot body material orientation. Kinematic parameters $\textcolor{black}{\textbf{v}}$ and $\textcolor{black}{\textbf{u}}$ describe geometric rates of change in linear and angular directions according to the local frame at $s$ (in Fig. \ref{fig:cross-section}(b), `x' and `y' are specifying the local coordinate system at $s$). 
\textcolor{black}{Furthermore, \textcolor{black}{the hat ( $\hat{}$ ) operator}  maps a vector in $\mathbb{R}^3$ to a skew-symmetric matrix in $SO(3)$ such that the cross product of two vectors can be represented as a matrix-vector multiplication.}
\textcolor{black}{The balance of \textcolor{black}{geometric linear and angular strain rates} of the continuum robot is formulated as below, which will be decomposed and discussed individually in the rest of this section.}
\textcolor{black}{
\begin{equation} \label{u_dot_v_dot}
    \begin{bmatrix}
        \dot{\textbf{v}} \\
        \dot{\textbf{u}}
    \end{bmatrix}=
    \begin{bmatrix}
        \textbf{K}_{\textbf{se}}+\textbf{K}_{\textbf{11}} && \textbf{K}_{\textbf{12}} \\
        \textbf{K}_{\textbf{21}} && \textbf{K}_{\textbf{bt}}+\textbf{K}_{\textbf{22}}
    \end{bmatrix}^{-1}
    \begin{bmatrix}
        \textbf{a} \\
        \textbf{b}
    \end{bmatrix}
\end{equation}
}
Stiffness matrix for shear and extension is denoted as $\textcolor{black}{\textbf{K}_{\textbf{se}}}=diag(GA,GA,EA)$, and similarly for bending and twist, $\textcolor{black}{\textbf{K}_{\textbf{bt}}}=diag(EI_x,EI_y,EI_z)$ is used. \textcolor{black}{$E$ is the elastic modulus, commonly put equal to 75GPa for nitinol,} and \textcolor{black}{similarly, $G$ is the shear modulus, commonly put equal to 25GPa for nitinol}. $A$ is the robot cross-section area, and \textcolor{black}{$I_x$, $I_y$, and $I_z$ are the second moments of area in the direction of $x$, $y$, and $z$, respectively}, all of which are constant over $s$. \textcolor{black}{Furthermore, $\textbf{K}_{\textbf{11}}=-\tau (\hat{\dot{\textbf{p}}}_\textbf{t}^\textbf{b})^2/\dot{||\textbf{p}_\textbf{t}^\textbf{b}}||^3$, where} $\tau$ is the applied tension to the actuation tendon, \textbf{$r$}=$[r_x, r_y,0]^T$ is the local representation of the tendon path, and $\textcolor{black}{\dot{\textbf{p}}_\textbf{t}^\textbf{b}=\hat{\textbf{u}}r+\dot{\textbf{r}}+\textbf{v}}$ is the tendon path unit tangent vector defined in the local frame. With $\textcolor{black}{\textbf{K}_{\textbf{11}}}$ obtained, $\textcolor{black}{\textbf{K}_{\textbf{12}} = -\textbf{K}_{\textbf{11}}\hat{\textbf{r}}}$, $\textcolor{black}{\textbf{K}_{\textbf{21}} = \hat{\textbf{r}}\textbf{K}_{\textbf{11}}}$, and $ \textcolor{black}{\textbf{K}_{\textbf{22}} = -\hat{\textbf{r}}\textbf{K}_{\textbf{11}}\hat{\textbf{r}}}$ are calculated consequently. Ultimately, the elements of the vector on the right\textcolor{black}{-}hand side of Eq. (\ref{u_dot_v_dot}) are defined.
\begin{equation} \label{eq:a-parameter}
\textcolor{black}{
    \textbf{a} = \textbf{K}_{\textbf{se}}\dot{\textbf{v}}^*-\hat{\textbf{u}}\textbf{K}_{\textbf{se}}(\textbf{v}-\textbf{v}^*)-\textbf{R}^T\textbf{f}_\textbf{e}-\textbf{K}_{\textbf{11}}(\hat{\textbf{u}}\dot{\textbf{p}}_\textbf{t}^\textbf{b}+\hat{\textbf{u}}\dot{\textbf{r}}+\ddot{\textbf{r}})
}
\end{equation}
\vspace{-25pt}
\begin{multline}
\textcolor{black}{
 \textbf{b} = \textbf{K}_{\textbf{bt}}\dot{\textbf{u}}^*-\hat{\textbf{u}}\textbf{K}_{\textbf{bt}}(\textbf{u}-\textbf{u}^*)-\hat{\textbf{u}}\textbf{K}_{\textbf{se}}(\textbf{v}-\textbf{v}^*)-}\\ \textcolor{black}{\textbf{R}^T\textbf{l}_\textbf{e}-\hat{\textbf{r}}\textbf{K}_{\textbf{11}}(\hat{\textbf{u}}\dot{\textbf{p}}_\textbf{t}^\textbf{b}+\hat{\textbf{u}}\dot{\textbf{r}}+\ddot{\textbf{r}})
}
\end{multline}
The \textcolor{black}{asterisk (}$^*$\textcolor{black}{)} notation indicates robot configuration when no tendon tension is applied. Additionally, \textcolor{black}{$\textbf{f}_\textbf{e}$} and \textcolor{black}{$\textbf{l}_\textbf{e}$} are the applied external distributed forces \textcolor{black}{(e.g. gravity)} and moments respectively.

\subsection{\textcolor{black}{Specific} Static Equilibrium Equations for \textit{ExoNav}} \label{subsec:design_and_model_2}

\begin{figure}[b]
    \centering
    \includegraphics[width=0.9\linewidth]{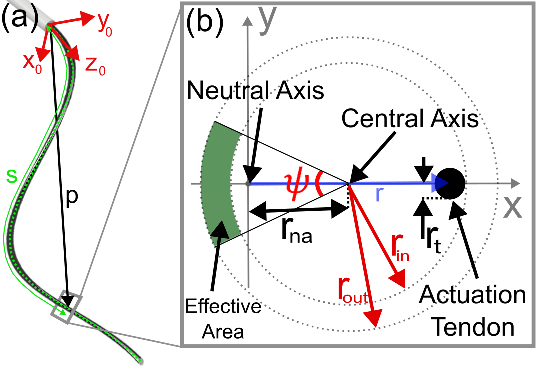}
    \caption{(a) \textit{ExoNav} bending into a helical curve, with the global coordinate frame defined at the robot base. The position vector $\textbf{p}(s)$ denotes the neutral axis at an arbitrary arc-length $s$. (b) Cross-sectional view at arc-length $s$, showing the locations of the actuation tendon, the neutral axis, and the effective cross-sectional area in the local coordinate frame.}
    \label{fig:cross-section}
\end{figure}

\begin{figure*}
    \centering
    \includegraphics[width=0.85\linewidth]{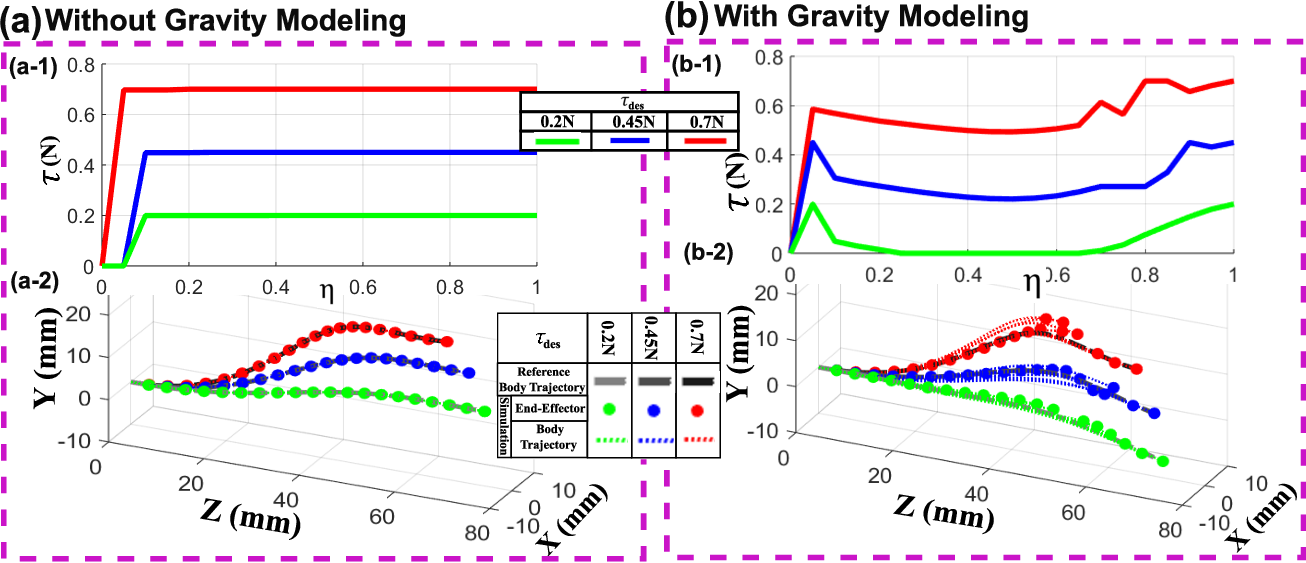}
    \caption{ \textcolor{black}{(a) ExoNav simulation results without gravity modeling: (a-1) optimized $\tau$ values along different FTL trajectories and (a-2) simulated ExoNav end-effector and body trajectories evolution compared to the reference FTL trajectory. (b) ExoNav simulation results with gravity effects included: (b-1) optimized $\tau$ values along different FTL trajectories and (b-2) simulated ExoNav end-effector and body trajectories evolution compared to the reference FTL trajectory.}}
    \label{fig:simulation}
\end{figure*}

To apply the Cosserat rod model to micromachined tube robots, we define the ``effective area'' (denoted by $A$) of the structure, which serves as the robot's backbone. In this context, the ``effective area'' refers to the unnotched, continuous longitudinal segment of the tube that appears as a circular arc, spanning the angle $\psi$ around the center, in each cross-section of the robot (see Fig. \ref{fig:cross-section}(b)). Accordingly, $\textcolor{black}{\textbf{p}}(s)$ is defined \textcolor{black}{pointed to} the neutral axis of this ``effective area'' \textcolor{black}{along the robot} (see Fig. \ref{fig:cross-section}(b)). \textcolor{black}{T}he rest of the circular cross-section of the tube can be ignored, since its sole purpose is to secure the actuation tendon inside the tube, not playing a role in robot deformations (as long as the edges of the neighboring notches do not contact due to high deformations).
\begin{equation}
    A = \frac{\psi (r_{out}^2-r_{in}^2)}{2}
\end{equation}
To define $\textcolor{black}{\textbf{p}}$, note that the neutral axis is deviated from the central axis by $r_{na}$\textcolor{black}{$=|(\int{xdA})/A|$} (See Fig. \ref{fig:cross-section}(b)).
\begin{equation}
    r_{na}=\frac{4}{3}\frac{sin(\psi/2)(r_{out}^3-r_{in}^3)}{\psi(r_{out}^2-r_{in}^2)}
\end{equation}
\textcolor{black}{Based on the neutral-axis deviation and the local coordinate frame origin, the material orientation matrix $\textbf{R}$ is defined such that the local x-axis always points toward the tube's central axis and away from the effective area (Fig. \ref{fig:cross-section}(b)). This definition makes the cross-sectional inertia independent of $s$, allowing the second moments of area, neutral-axis location, and tendon position to remain constant along the robot. An alternative formulation with constant $\textbf{R}$ and varying cross-sectional inertia results in significantly more complex governing equations with additional derivative terms. The second moments of area about the x- and y-axes are then obtained at the neutral axis using the parallel axis theorem, following a similar approach to \cite{jeong2020design}.}

\begin{equation}
    I_x = \frac{(\psi+sin(\psi))(r_{out}^4-r_{in}^4)}{8}
\end{equation}
\begin{equation}
    I_y = \frac{(\psi-sin(\psi))(r_{out}^4-r_{in}^4)}{8}-Ar_{na}^2
\end{equation}
And subsequently, $I_z=I_x+I_y$.

\textcolor{black}{Assuming the actuation tendon follows the shortest available path under tension, it occupies a position diametrically opposite to the effective area in the micromachined tube (Fig. \ref{fig:cross-section}(b)). Accordingly, the local tendon position is defined as $\textbf{r}=[r_{na}+r_{in}-r_t,0,0]^T$. The local frame at $s=0$ (robot base) is aligned with the global frame (Fig. \ref{fig:cross-section}(a)). A key aspect of the ExoNav model is the definition of the initial backbone position $\textbf{p}^*$. Because the backbone is represented by the effective cross-sectional area, the initial position is defined along the local neutral axis, which traces a slight helical path along $s$. This backbone definition, together with the nontrivial initial orientation matrix $\textbf{R}^*$, distinguishes ExoNav from previously reported robots with helical tendon paths \cite{rucker2011statics, kheradmand2025comprehensive}.}
\begin{equation} \label{p_star}
    \textcolor{black}{\textbf{p}}^* = \begin{bmatrix}
        -r_{na}cos(2\pi s/L_{na}) \\
        -r_{na}sin(2\pi s/L_{na}) \\
        Ls/L_{na}
    \end{bmatrix}
\end{equation}
\begin{equation} \label{R_star}
    \textcolor{black}{\textbf{R}}^*=\begin{bmatrix}
        cos(2\pi s/L_{na}) && -sin(2\pi s/L_{na}) && 0 \\
        sin(2\pi s/L_{na}) && cos(2\pi s/L_{na}) && 0 \\
        0 && 0 && 1
    \end{bmatrix}
\end{equation}
\textcolor{black}{T}he length of the neutral axis is denoted by $L_{na}=\sqrt{L^2+(2\pi r_{na})^2}$. Substituting $\textcolor{black}{\textbf{p}}^*$ and $\textcolor{black}{\textbf{R}}^*$ in Eqs. (\ref{p_dot}) and (\ref{R_dot}), $\textcolor{black}{\textbf{v}}^*$ and $\textcolor{black}{\textbf{u}}^*$ are obtained.
\begin{equation}
    \textcolor{black}{\textbf{v}}^*=[0, -2\pi r_{na}/L_{na}, L/L_{na}]^T
\end{equation}
\begin{equation}
    \textcolor{black}{\textbf{u}}^*=[0,0,2\pi/L_{na}]^T
\end{equation}

By substituting these specific parameters in Eqs. (\ref{p_dot}), (\ref{R_dot}), and (\ref{u_dot_v_dot}), the overall model equations are obtained. 
For state variables $\textcolor{black}{\textbf{p}}$ and $\textcolor{black}{\textbf{R}}$ in the model equations, initial conditions are obtained by substituting $s=0$ in Eqs. (\ref{p_star}) and (\ref{R_star}). Furthermore, tendon termination boundary conditions are defined for the kinematic state variables $\textcolor{black}{\textbf{v}}$ and $\textcolor{black}{\textbf{u}}$.
\begin{equation}
    \textcolor{black}{\textbf{v}(L_{na}) = \textbf{v}^*(L_{na})-\textbf{K}_{\textbf{se}}^{-1}(\tau \frac{\dot{\textbf{p}}_\textbf{t}^\textbf{b}(L_{na})}{\norm{\dot{\textbf{p}}_\textbf{t}^\textbf{b}(L_{na})}})}
\end{equation}

\begin{equation}
    \textcolor{black}{\textbf{u}(L_{na}) = \textbf{u}^*(L_{na})-\textbf{K}_{\textbf{bt}}^{-1}\hat{\textbf{r}}(L_{na})(\tau \frac{\dot{\textbf{p}}_\textbf{t}^\textbf{b}(L_{na})}{\norm{\dot{\textbf{p}}_\textbf{t}^\textbf{b}(L_{na})}})}
\end{equation}
\textcolor{black}{where the actuation tendon tension $\tau$ is the model input. The governing equations are implemented in MATLAB using the ode45 solver and solved with a shooting method based on fsolve. Initial guesses for $u$ and $v$, together with the known initial position and orientation, are integrated along the robot. The resulting terminal values of $u$ and $v$ are then compared with the prescribed end conditions, and the initial guesses are iteratively updated until the residual falls below a specified tolerance, yielding the robot configuration.
}

\subsection{Progressive and FTL Motion \textcolor{black}{Simulation}} \label{subsec:design_and_model_3}

We define ``progressive motion'' as a kinematic behavior primarily characterized by translational movement along the robot's longitudinal axis, which may be accompanied by rotation. Let $L_p$ denote the exposed length of the robot extending beyond an enclosing outer tube. To quantify the extent of this motion, we introduce the ``progression factor'' $\eta \in [0,1]$, where $\eta = 0$ corresponds to the robot being fully retracted, and $\eta = 1$ represents the robot being fully extended out.
\begin{equation}
    L_p = \eta L
\end{equation}
Note that a necessary condition for classifying a motion as ``progressive motion'' is a change in the progression factor $\eta$ (either increasing or decreasing). Within this context, the arc-length parameter in ``progressive motion'' is defined as $s_p \in [0, \eta L_{na}]$ \textcolor{black}{(focused only on the exposed length of the robot)}. Furthermore, the tip of the outer tube is designated as the robot base, as it marks the onset of deformation along the robot body. Consequently, the initial values $\textcolor{black}{\textbf{p}}(0)$ and $\textcolor{black}{\textbf{R}}(0)$ are redefined by substituting $s = (1 - \eta)\textcolor{black}{L_{na}}$ into Eqs. (\ref{p_star}) and (\ref{R_star}), thereby accounting for the new base location.

A helical deformable structure is naturally capable of performing FTL motion upon a simultaneous rotation and translation (related through the pitch of the helix, which is $\frac{L}{2\pi}$ for the case of \textit{ExoNav}) as it comes out of a rigid outer tube. In this regard, the FTL motion is a subset of the general ``progressive motion''. In our case, the FTL motion requirement depends on the applied rotation ($\phi_p= 2\pi(\eta-1)$) and tendon tension ($\tau$) along the progressive motion. \textcolor{black}{The applied tendon tension can be programmed and pre-adjusted for a specific helical shape for the desired FTL trajectory.} A ``progressive motion'' simulation is utilized \textcolor{black}{based on the developed Cosserat rod statics model, }to obtain the appropriate tendon tension input. 

\begin{figure*}
    \centering
    \includegraphics[width=0.85\linewidth]{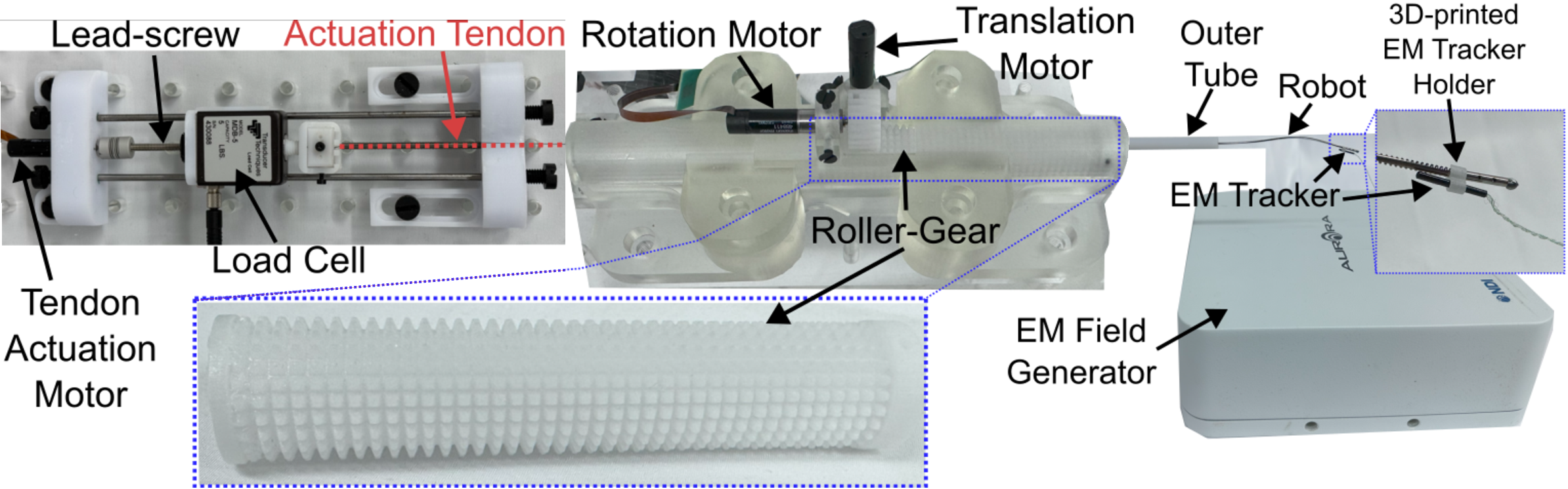}
    \caption{Components of the actuation setup, with the tendon pulling mechanism on the left, the dual-DoF roller-gear mechanism in the center (inset shows circumferential and longitudinal gear heads), and the robot protruding from the outer tube on the right. An EM sensor is attached to the robot via a 3D-printed holder (inset on the right), with the EM field generator positioned beneath the exposed robot.}
    \label{fig:setup}
\end{figure*}

\textcolor{black}{To simulate FTL motion, the model is first solved with $\eta=1$ and a desired tendon tension $\tau_{des}$, whose resulting robot shape defines the FTL reference trajectory. The simulation then progresses by incrementing $\eta$ from 0 to 1 with step size $\Delta\eta$, while solving the model for different $\tau$ values at each iteration until the exposed robot body sufficiently matches the reference trajectory. During this process, $L_p$ and $\phi_p$ are updated to satisfy the FTL motion requirements. Simulations were performed for $\tau_{des}=0.2$, 0.45, and 0.7 N with $\Delta\eta=0.05$. In all cases, the optimized tendon tension remained essentially constant throughout deployment (Fig. \ref{fig:simulation}(a-1)), while the robot body and tip closely followed the reference trajectory (Fig. \ref{fig:simulation}(a-2)). These results indicate that FTL navigation can be achieved by maintaining a constant tendon tension as the robot is translated and rotated out of the outer tube. For smaller $\tau_{des}$, the optimized tension may initially remain at 0 N because the partially exposed straight robot already approximates the reference trajectory. Although this simulation neglects external forces, the optimization framework can also determine tendon tensions that minimize deviations from the reference trajectory when such forces are present.
}

\subsection{\textcolor{black}{Gravity Effect Modeling}} \label{subsec:design_and_model_4}

The exerted gravity force to the robot can be calculated and inserted into the model as a uniformly distributed external force ($\textcolor{black}{\textbf{f}_\textbf{e}}$ in Eq. \ref{eq:a-parameter}). To do so, the mass density of nitinol ($\rho=$6255$Kg/m^3$) is utilized to calculate the linear mass density ($\lambda$) of ExoNav, considering the notches along the robot:

\begin{equation}
    \lambda = \frac{(\pi d + \psi w /2)(r_{out}^2-r_{in}^2)}{d+w} \rho
\end{equation}
Assuming the gravity force being towards $-X_0$ (according to the depicted global frame in Fig. \ref{fig:cross-section}(a)), the external gravitational force can be calculated as:
\begin{equation}
    \textcolor{black}{\textbf{f}_\textbf{e}}=\lambda g [-1,0,0]^T 
\end{equation}
where $g=$~9.81 $m/s^2$ is the gravitational acceleration.

\textcolor{black}{This external force is implemented in the simulation and model equations, and a similar simulation instruction (as described in subsection \ref{subsec:design_and_model_3}) is followed. The optimized $\tau$ for each simulation with a specific $\tau_{des}$ seem completely different from the prior results where no gravity effect was modeled. In the first iterations, $\tau$ increases to a certain value commonly lower than its corresponding $\tau_{des}$, which then follows a slight U-shaped curve, and in final iterations, optimized $\tau$ values seem to converge to $\tau_{des}$ (See Fig. \ref{fig:simulation}(b-1)). FTL reference trajectories and simulated robot body trajectories show significant bending towards $-X_0$ (gravitational force direction) compared to the case with no gravity modeling (See Fig. \ref{fig:simulation}(a-2) and (b-2)). The robot body trajectories and end-effector along the simulated progressive motion are kept at a close distance from the FTL reference trajectory, with a few observable deviations, especially towards the end of the progressive motion (See Fig. \ref{fig:simulation}(b-2)). These deviations appear as some jaggedness in the optimized $\tau$ values (See Fig. \ref{fig:simulation}(b-1)). These deviations and jaggedness are most probably due to relatively high error margins or multiple local minima in the search span of the simulation program. There are several ways to resolve these issues to have closer alignments with the FTL reference trajectory and subsequently, smoother optimized $\tau$ values over $\eta$ variations, which are potential focuses for future studies.
}

\begin{figure}
    \centering
    \includegraphics[width=0.80\linewidth]{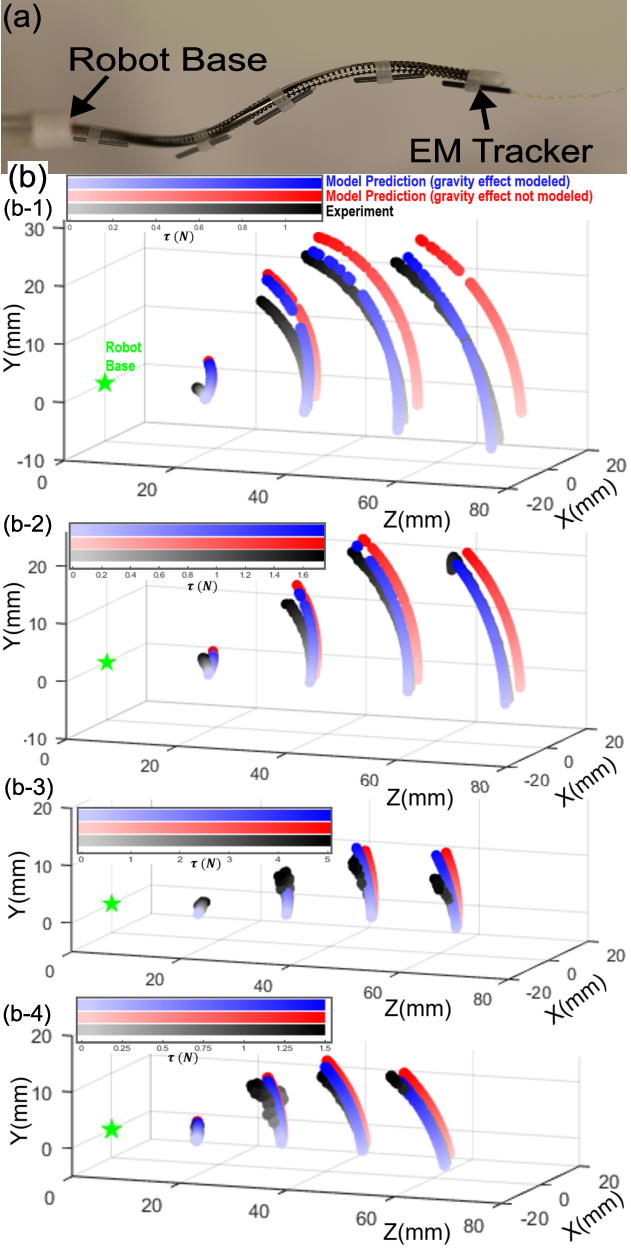}
    \caption{(a) Robot fully extended from the outer tube and bent into a helical shape, with the EM sensor placed at four distinct locations along its length and at the base. (b) Comparison of experimental position measurements with model predictions\textcolor{black}{, with and without gravity modeling,} for (b-1) prototype 1, (b-2) prototype 2, (b-3) prototype 3, and (b-4) prototype 4.}
    \label{fig:validation_results}
\end{figure}

\section{EXPERIMENTAL RESULTS} \label{sec:simulation_and_experiment}

\subsection{Experimental Setup} \label{subsec:simulation_and_experiment_1}

The experimental setup consists of a motorized lead-screw mechanism capable of linearly translating a bracket, which holds a load-cell (MDB-5, Transducer Techniques) to measure tendon tensions (see Fig. \ref{fig:setup}). To provide translation and rotation DoFs, we used a motorized roller-gear mechanism \textcolor{black}{\cite{morimoto2017design}}, capable of simultaneous rotation and translation. Additionally, 3D-printed outer tubes are customized for each prototype (see Fig. \ref{fig:setup}). We utilized an electromagnetic (EM) tracking system (NDI Aurora, Northern Digital Inc, Waterloo, Ontario, Canada) to measure the position of different points on the robot. EM tracking system consists of two main parts: EM field generator and a small tracking module, called the EM sensor. Small components were 3D-printed to provide connection between the EM sensor and the robot body (see Fig. \ref{fig:setup}). \textcolor{black}{The offset between the EM sensor and the robot body is measured and compensated, hence all the provided results are denoting the measured points on the robot body.} Three DC motors (RE 8 Ø8 mm, Precious Metal Brushes, 0.5 Watt, with terminals, Maxon international ltd., Switzerland) are used in tendon pulling and roller-gear mechanisms, all of which are controlled by a Texas Instrument LAUNCHXL-F28379D LaunchPad, utilized for motor control and to gather force and motor encoder data in the MATLAB/Simulink environment. Furthermore, we gathered the synchronized position data from EM tracking system in MATLAB.

\begin{table}[b]
\caption{$p(L_{na})$ estimation errors with and without gravity included.}
\label{tab:errros}
\begin{center}
\begin{tabular}{|c|c|c|c|c|c|}
\hline
  \multicolumn{2}{|c|}{Prototypes} & 1 & 2 & 3 & 4 \\
\hline
\hline
Gravity & MED~(mm) & 7.71 & 6.93 & 11.03 & 6.92 \\
Not Included & RMSE~(mm) & 4.18 & 3.30 & 9.60 & 5.61 \\
\hline
\hline
Gravity & MED~(mm) & 4.32 & 5.03 & 5.76 & 2.43 \\
Included & RMSE~(mm) & 1.76 & 2.33 & 2.18 & 1.33 \\
\hline
\end{tabular}
\end{center}
\end{table}

\subsection{Model Validation \textcolor{black}{Experiment}} \label{subsec:simulation_and_experiment_2}

\begin{figure*}
    \centering
    \includegraphics[width=1.0\linewidth]{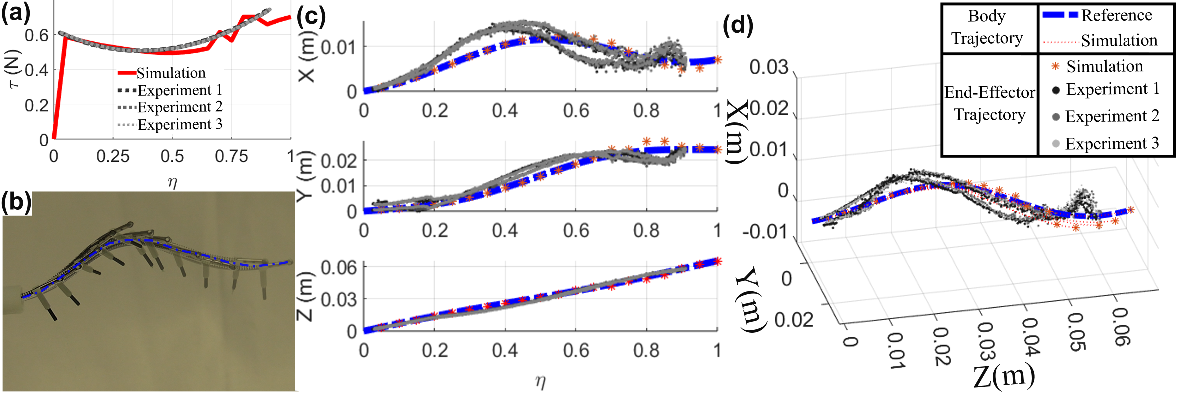}
    \caption{(a) Simulated and experimental tendon tensions plotted as a function of the progression factor during FTL motion. (b) Robot body motion in the FTL experiment closely follows the reference FTL trajectory. (c) End-effector position plotted as a function of progression factor in simulation and experiments compared to the reference trajectory. (d) 3D visualization of end-effector positions from simulation and experiments alongside the reference FTL trajectory.}
    \label{fig:ftl_results}
\end{figure*}
\begin{figure}[t]
    \centering
    \includegraphics[width=1.0\linewidth]{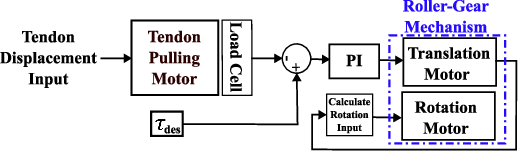}
    \caption{Schematic of the tendon tension control system implemented during the FTL motion experiment.}
    \label{fig:control}
\end{figure}
\begin{figure}[b]
    \centering
    \includegraphics[width=1.0\columnwidth]{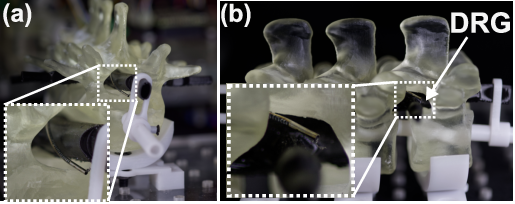}
    \caption{\textcolor{black}{ExoNav navigation to the (a) ventral spinal cord and (b) DRG region inside a spinal cord phantom model.}}
    \label{fig:demo}
\end{figure}

Conducting experimental trials on each prototype, first, the EM sensor is placed at the robot's base to set the origin of the global frame. Then the EM sensor is placed at locations closely corresponding to arc-length values of $s=$0.25$L_{na}$, $s=$0.5$L_{na}$, $s=$0.75$L_{na}$, and $s=L_{na}$, and the robot is actuated in each case, \textcolor{black}{with specific tendon tension ranges for each prototype (See Fig. \ref{fig:validation_results}(b)).} \textcolor{black}{For each sensor placement, the robot is actuated, resulting in four distinct bending trials per prototype, with the EM sensor location updated between trials} (see Fig. \ref{fig:validation_results}(a)). \textcolor{black}{Similar experiments are conducted across different prototypes, and $\tau$ values as well as position data are recorded. The gathered $\tau$ values are then fed into the developed model to predict ExoNav's deformations with and without gravity modeling (See Fig. \ref{fig:validation_results}(b)). The model prediction results show close alignment between the experiment data and the gravity-included model predictions, while model predictions without including gravity are deviated (especially in more flexible prototypes, e.g. prototype 1 shown in Fig. \ref{fig:validation_results}(b-1)) from the experiment data, emphasizing the significant effect of external forces on the resulted robot deformation and the necessity to model these external loads. In addition to the obtained graphs (in Fig. \ref{fig:validation_results}(b)),} corresponding maximum \textcolor{black}{Euclidean} distance \textcolor{black}{(MED)} errors and \textcolor{black}{root mean square errors (RMSE)} \textcolor{black}{of end-effector position estimations} are reported \textcolor{black}{for predictions with and without modeling gravity} (see Table \ref{tab:errros}). \textcolor{black}{These results show that, overall, including gravity effects in the model reduces the MED and RMSE values.}

\subsection{Tendon Tension Control \textcolor{black}{and FTL Motion Experiment}} \label{subsec:simulation_and_experiment_3}

To perform an arbitrary ``progressive motion'', three input commands, including rotation, translation, and tendon tension, must be applied to the experimental setup. However, for FTL motion, rotation and translation inputs are obtained in a dependent fashion, based on $\eta$. Therefore, a feedback control system with a PI controller is implemented in Simulink and connected to the utilized micro-controller board. Considering the low speed of the tendon pulling mechanism compared to the motors of the roller-gear mechanism, the controller is implemented on the DoFs that the roller-gear mechanism provides, limiting the overall actuation velocity to the motor of the tendon pulling mechanism, hence applying $\eta$ indirectly. In this tendon tension control system, we apply a tendon displacement value to the tendon pulling mechanism, which causes the tendon tension to change, followed by the roller-gear translation motor \textcolor{black}{actuation based on the PI controller output}, so that the tendon tension gets adjusted. Based on the applied translation, $\eta$ is calculated, and the appropriate rotation is obtained consequently, and is applied to the rotation motor of the roller-gear mechanism (see Fig. \ref{fig:control}). We iteratively tested and finally adjusted the PI controller proportional and integral gains as $k_P=k_I=$~50.

As observed in \textcolor{black}{subsection \ref{subsec:design_and_model_4}}, the gravity-induced deformations make the FTL reference trajectory deviate to a \textcolor{black}{distorted} helix \textcolor{black}{which requires a non-constant $\tau$ along $\eta$ to keep ExoNav close to the corresponding FTL reference trajectory. To avoid sudden changes in the actuation mechanism due to jaggedness observed in $\tau$ signals (See Fig. \ref{fig:simulation}(b-1)), }a second-order polynomial is fitted \textcolor{black}{(using MATLAB curve fitting toolbox)} to the \textcolor{black}{optimized $\tau$ signal for $\tau_{des}=$0.7N} (see Fig. \ref{fig:ftl_results}(a)). \textcolor{black}{The second-order polynomial curve, $\tau_{FTL}(\eta)=0.6492 \eta^2 - \textcolor{black}{0.}5159\eta + 0.6088$, is utilized to conduct three trials of FTL motion experiment using prototype 1.}

The EM tracking module is placed at the tip of the robot to gather end-effector position data (see Fig. \ref{fig:ftl_results}(b)). The obtained data from both simulation and all three experiments are compared to the FTL reference trajectory (see Fig. \ref{fig:ftl_results}(c) and (d)). The simulation results show close alignment with the reference trajectory and an RMSE value of 1.0706~mm. The experiment results show RMSE values of 3.6433~mm, 3.6288~mm, and 3.7481~mm for experiment \textcolor{black}{trials} 1 to 3, respectively. Comparison between different experiment trials shows a consistent \textcolor{black}{and repeatable} behavior. End-effector position errors in these experimental trials are primarily due to frictions (between the outer tube inner walls and the robot), EM tracking module assembly weight, and tendon tension polynomial approximation errors. \textcolor{black}{Note that the reported end-effector position data focus solely on the end-effector trajectory as a representative sample of the full robot body executing a FTL motion. Nevertheless, visual inspection qualitatively confirms that the entire body undergoes FTL motion (see Fig. \ref{fig:ftl_results}(b)).}

\textcolor{black}{ \subsection{Robot Demonstration in Phantom Spinal Cord Model} \label{subsec:robot_pahntom_demo}}

\textcolor{black}{To demonstrate ExoNav's ability to reach clinically relevant anatomical targets, we conducted a set of navigation experiments inside a spinal cord phantom model which includes a 3D-printed mock spinal cord placed inside the 3D-printed spinal canal. The robot was inserted through the dorsal entry and a joystick was used to provide teleoperation for the operator, controlling translation, rotation, and tendon actuation via dedicated joystick key pairs.} \textcolor{black}{The operator successfully navigated to the lateral and then ventral target areas (See Fig. \ref{fig:demo}(a)). The navigation process consisted of iterative translations and tendon actuation, which was then accompanied by corrective rotations to adjust the helical shape of ExoNav to reach the targets. The user also attempted to navigate to the dorsal root ganglia (DRG) regions and successfully navigated to these additional targets (See Fig. \ref{fig:demo}(b)). Navigating to the DRGs is a promising capability for pain management applications which was not considered as a primary goal of ExoNav design, but it was found capable to achieve it. Overall, the user was able to navigate around the spinal cord with ease, avoiding unnecessary maneuvers that are common with manual SCS navigation techniques. Although ExoNav is manually controlled in these demonstrations, the observed paths are qualitatively consistent with the proposed model and FTL motion paths. ExoNav does not include an SCS electrode integration mechanism and exploring possible latching mechanisms for this purpose is one of the future horizons for this research line.}

\section{CONCLUSION} \label{sec:conclusion}
\textcolor{black}{In this manuscript, we described the design of ExoNav, a notched-tube continuum robot with convoluted notch patterns to acquire helical paths to reach lateral and ventral spinal cord fibers to implement SCS electrodes in proximity to the motor tracts. We applied Cosserat rod model equations for tendon-driven continuum robots to include the convoluted notch design and implemented a simulation environment to validate the developed model and analyze ExoNav's capability to execute FTL motion. The effect of gravity as an example of a possible external force is investigated and implemented in the developed model. We also validated our model and tested ExoNav FTL operation experimentally, using an actuation tendon tension PI control system. Ultimately, we show ExoNav capability in maneuvering to the lateral and ventral spinal cord tracts, as well as the DRG of a phantom spinal cord model. Future direction include integration of the actual SCS electrode on the robot and further developing the tendon tension optimization program for real-time applications.}

\bibliographystyle{IEEEtran}
\bibliography{references}
\end{document}